\documentclass[conference]{IEEEtran}
\IEEEoverridecommandlockouts
\usepackage{cite}
\usepackage{amsmath,amssymb,amsfonts}
\usepackage{algorithmic}
\usepackage{graphicx}
\usepackage{textcomp}
\usepackage{xcolor}
\usepackage{float} 
\usepackage{booktabs}
\usepackage{tabularx}
\usepackage{array}
\usepackage[utf8]{inputenc}
\usepackage{url} 

\newcolumntype{C}{>{\centering\arraybackslash}X}

\def\BibTeX{{\rm B\kern-.05em{\sc i\kern-.025em b}\kern-.08em
    T\kern-.1667em\lower.7ex\hbox{E}\kern-.125emX}}
\begin{document}

\title{Automated Analysis of Learning Outcomes and Exam Questions Based on Bloom's Taxonomy}

\author{
\IEEEauthorblockN{Ramya Kumar$^{1}$, Dhruv Gulwani$^{1}$, Sonit Singh$^{1}$}
\IEEEauthorblockA{\textit{$^{1}$School of Computer Science and Engineering, University of New South Wales, Sydney, Australia}\\
\textit{Email: ramya.kumar1@unsw.edu.au}
}}

\maketitle

\begin{abstract}
This paper explores the automatic classification of exam questions and learning outcomes according to the Bloom’s Taxonomy. A small dataset (600 sentences) labelled with six cognitive categories - Knowledge, Comprehension, Application, Analysis, Synthesis, and Evaluation was processed using traditional machine‐learning (ML) models (Naïve Bayes, Logistic Regression, Support Vector Machines), Recurrent Neural Network architectures (LSTM, BiLSTM, GRU, BiGRU), Transformer‐based models (BERT and RoBERTa), and Large Language Models (OpenAI, Gemini, Ollama, Anthropic). Each model was evaluated under different preprocessing and augmentation strategies (e.g., synonym replacement, word embeddings, etc.). Among traditional ML approaches, Support Vector Machines (SVM) with data augmentation achieved the best overall performance, reaching 94\% accuracy, recall, and F1‐scores, with minimal overfitting. In contrast, the RNN models and BERT suffered from severe overfitting. RoBERTa overcame overfitting initially but eventually began to show signs of it as training progressed. Finally, zero‐shot calls to large language models (LLMs) indicated that OpenAI and Gemini performed best among the tested LLMs with ~0.72–0.73 accuracy and commensurate F1‐scores. These findings highlight the challenges of training complex deep models on limited data and underscore the value of careful data augmentation and simpler algorithms (like augmented SVM) for Bloom’s Taxonomy classification.
\end{abstract}


\section{Introduction}
The evaluation and classification of exam items based on the Taxonomy of Bloom~\cite{forehand:2005:bloom} has been a growing issue of concern in educational data mining as well as in the design of instruction. Having been first introduced as a hierarchical model of the systematic grouping of cognitive abilities, the Taxonomy suggested by Bloom offers an ordered sequence of the development of cognitive skills by beginning with the ability to recollect the existing knowledge and continuing with the development of higher levels of analysis and evaluation. Mapping learning resources and test questions to these six levels of cognitive processing, educators can both create more balanced learning programs and also match the difficulty involved in assessments to the desired outcomes of that learning, as well as assure constructive fit across a course or program. In addition to metacognitive awareness and more effective preparation strategies, students can have a better understanding of the cognitive demands of every question.

Although it has pedagogical importance, manually categorising the exam questions or learning outcomes by the levels of Bloom is a very tedious, labour intensive and error-prone task. Within mass education systems, \emph{e.g.} Massive Open Online Courses (MOOCs), institutional assessment banks, or program-wide accreditation exercises, the number of items that need to be classified surpasses the ability of human to manually classify them. Also, there is an issue of inconsistency because human evaluation is subjective bringing inter-rater variability, in turn compromising the reliability of application of Bloom's taxonomy to classify questions or learning outcomes. These limitations emphasised the importance of having a scalable and automated system that could provide high level of accuracy and consistency in performing Bloom-level classification.

The recent developments in Natural Language Processing (NLP) have showed significant progress in automated text understanding and generation. Although there has been work on automated classification of questions and learning outcomes based on the Bloom's taxonomy \cite{Li:2022:automatic_classification, Huang:2022:automatic_classroom, ALMATRAFI:2025:GenAI, waheed-etal-2021-bloomnet}, several research gaps remain unsolved. First, most of the existing studies use conventional text features in combination of classical machine learning algorithms. Definitely, there is a need to have systematic evaluation of conventional machine learning methods and more recent deep learning methods. Second, with the recent developments in Large Language Models (LLMs), it opens up new possibilities of Zero-Shot or Few-Shot classification. Third, there is a need to check the effectiveness of deep learning methods on small-size datasets, as annotating large-scale dataset is time consuming, costly, and laborious. 

In this paper, we build and test an end-to-end pipeline to automate the process of classification of the exam questions and learning outcomes across all the six levels of cognition as per Bloom's taxonomy. We compared the performance of several modelling paradigms with a small dataset of 600 labelled sentences: traditional machine-learning classifiers (Naïve Bayes \cite{ML_book}, Logistic Regression \cite{ML_book, Logistic_Regression}, Support Vector Machines (SVM) \cite{SVM}), recurrent neural networks (RNNs) (LSTM \cite{LSTM}, BiLSTM, GRU \cite{GRU}, BiGRU), transformer-based models (BERT \cite{devlin-etal-2019-bert} and RoBERTa \cite{Liu:Roberta}), and state-of-the-art large language models (OpenAI GPT-4o-mini \cite{Brown2020gpt3, Ouyang2022instructgpt}, Google Gemini-1.5-Pro \cite{geminiteam2023gemini, Imran2024}, LLaMA 3.1 \cite{touvron:2023:llama, llama3} and Anthropic Claude-3.5-Haiku \cite{claude3}). To address the issue of small-size dataset, we use synonym augmentation, and test the effects of pretrained GloVe embeddings \cite{pennington-etal-2014-glove} to determine their effect on generalisation. Lastly, we assess LLMs in zero-shot to explore their applicability as training-free, lightweight alternatives. In general, our work is a thorough comparative study of the current NLP methods of Bloom’s Taxonomy classification based on the realistically small-size data. The results reveal the unexpectedly strong competitiveness of traditional ML models, the overfitting vulnerability of RNNs and some transformers, and the potential of LLMs despite the absence of fine-tuning. These lessons add to the current research on the automation of education and shape further work to develop reliable, scalable, and pedagogically grounded assessment-analysis systems.

\section{Related Work}
In this section, we present the description of the Taxonomy of Bloom and discuss the literature devoted to the automation of the process of analysis and classification of learning outcomes and examination questions with the help of this cognitive structure. The Taxonomy developed by Bloom has been widely used as a model of explaining cognitive learning goals, which provides a hierarchical structure of the most basic knowledge recollection processes to the most advanced skills of analysis, synthesis, and evaluation.

With the growing use of educational institutions in the data-driven assessment practice, the necessity of automatically categorising questions by such levels of cognition has become a central issue in educational data mining and NLP. The automation of Bloom-level classification has thus become an increasingly popular field of research, and initial work on the topic has been largely based on machine-learning (ML) and feature-engineering methods. The hybrid approaches suggested by Mohammed and Omar \cite{mohammed2019} include the combination of Term Frequency with Parts-of-Speech  - Inverted Document Frequency (TFPOS-IDF), a TF-IDF representation with a part of speech information weighted and Word2Vec embeddings as one of the basic hybrid methods, which classify the questions into levels of Bloom Cognition. Their findings indicated that adding POS-weighted feature and semantic representation of vectors (verbs as a signature of cognitive actions) as well as semantic meaning, are critical in improving classification accuracy in terms of syntactic cues (\emph{e.g.}, verbs as a signature of cognitive actions) and semantic meaning, which is of vital importance in distinguishing between the Bloom categories. Another study by Harrison \emph{et al.,}~\cite{harrison2019}, created a classifier based on revised taxonomy developed by Bloom. Authors used a combined dataset, which contained pre-labelled samples and a manually annotated corpus of over 1,500 items so that they could effectively evaluate it. By conducting extensive feature engineering on the short text of the educational texts, they compared the different traditional ML algorithms and showed which models worked best with different linguistic representations. Good quality of annotation and the relevance of linguistic clues in particular domains were also highlighted in this study as far as the development of automated classifiers in educational material is concerned.

The question complexity was considered in a complementary viewpoint presented by Ullrich and Geierhos \cite{ullrich2019}, who investigated it beyond the surface characteristics. They specialised in the task of differentiating between the simple fact-based questions and those that involve multi-hop or multi-step reasoning. Authors showed that the models should follow more contextual knowledge and not only the recognition of the keywords by analysing linguistic structures and contextual dependencies in order to classify cognitive complexity. This points to a significant weakness of the previous ML-based methods that frequently had problems with subtle differences between higher-order Bloom levels. Recent studies have shifted in the direction of deep learning and transformer-based models due to the constraints of traditional feature-engineered systems. Li \emph{et al.} \cite{li2020} developed one of the large dataset, consisting of 21,380 learning objectives of over 5,500 courses. In their comparative study, they had to compare classical ML models (e.g., Naive Bayes, Logistic Regression, SVM, Random Forest, XGBoost) with the performance of BERT-based classifiers and found that pretrained language models are more successful, especially when distinguishing between finer levels of cognition. This observation is indicative of the wider NLP trend that is in support of contextualised embeddings instead of engineered features.

Furthermore, RyanLauQF \cite{bloombert2022} proposed the BloomBERT transformer-based classifier, which is specifically intended to classify the tasks related to productivity through the Bloom cognitive framework. BloomBERT has API-native deployment, making it possible to directly integrate with real-life learning management systems and educational tools. It demonstrates how transformer architectures may be realised in production-ready scalable environments, with automated cognitive classification enabling curriculum design, feedback creation and adaptive learning systems.

In summary, literature review exemplify how the research on the automated classification of questions based on Bloom's taxonomy has progressed: rule-based to feature-based ML methods, then, recurrent neural networks and, transformer-based and recently, LLM-based. Literature also point out that there are problems that are persistent, including few labelled datasets, overfitting in deep models, domain-specific ambiguity, and interpretations of model decisions. To deal with such issues, it will continue to be an important direction in the development of the reliable, scalable automated systems able to perform cognitive-level classification in education.

\section{Methodology}
Our proposed pipeline consists of three experimental stages:

\begin{enumerate}
    \item Traditional ML methods
    \item RNN-based methods
    \item Transformer-based methods
    \item LLM-based methods
\end{enumerate}

The data was composed of 600 labelled sentences containing 100 sentences of each of the labels of Bloom: Knowledge, Comprehension, Application, Analysis, Evaluation, and Synthesis. 

\subsection{Data preprocessing}
There were two elements in each dataset entry: a Sentence (exam question or learning outcome) and a Label (one of the six Bloom categories). In order to guarantee consistency and quality before modelling, we preprocessed them using the following steps:

\begin{itemize} 
\item lowercasing all text, 
\item elimination of punctuation and non-alphanumeric characters using regular expressions, step word-level tokenisation of sentences, 
\item work out lemmata with the lemmatiser in WordNet,
\item stop word removal by the NLTK stopword list,
\item filtering tokens of length no more than $\leq 2$. 
\end{itemize}  

During preprocessing, we removed sentences having no text. The processed data was factorised into numerical class labels and stratified split into training and test sets with a 80/20 stratified split. 

\subsection{Data augmentation}
Given the small-size of the dataset and to avoid model overfitting problem, we applied data augmentation technique, namely, \emph{synonym‐replacement} as given the following steps: 

\begin{enumerate}
    \item identify eligible words in the sentence (words not in stopwords, length $> 3$)
    \item select a random word and replace it with a randomly chosen synonym from WordNet. 
\end{enumerate}

The data augmentation was performed at a rate of $\approx 10\%$ of the original data in most experiments. 

\subsection{Machine Learning methods}
We selected classical ML methods, namely, Naive Bayes (NB), Logistic Regression (LR) and Support Vector Machines (SVM), trained and evaluated on three different feature setups. The Bag-of-Words \emph{countVectorizer} was used to represent the text and part-of-speech (POS) features. The Synthetic Minority Oversampling Technique (SMOTE) was used during training in order to cope with small imbalance in classes.  Each of the ML method was tuned with the use of \emph{GridSearchCV} with five-fold cross-validation on the training split. We used standard classification metrics, namely, \emph{accuracy}, \emph{precision}, \emph{recall}, \emph{Micro F1}, and \emph{Macro F1} to evaluate model performance. We also checked performance on the training to check if there are any signs of model overfitting.

\subsection{Dep Learning methods}

\subsubsection{Recurrent Neural Networks (RNNS)}

We applied four robust RNNs, namely, Long Short-Term Memory (LSTM) \cite{LSTM}, Bi-directional LSTM (Bi-LSTM), Gated Recurrent Unit (GRU) \cite{GRU}, and Bi-directional GRU (Bi-GRU).  In order to process the raw text, we tokenised it using Keras’s Tokenizer ($num\_words=10000$), and sequences were padded to a maximum length of $100$, and GloVe embeddings ($300$‐dimensional) were loaded. We tested the performance of four RNNs under the following conditions: 

\begin{enumerate}
    \item Model alone (basic preprocessing)
    \item Model + Data Augmentation (synonym replacement)
    \item Model + Embeddings (GloVe 42B.300d)
    \item Model + Embeddings + Data Augmentation
\end{enumerate}

Each of these models were trained and tested independently. To avoid overfitting, we use early stopping and set variable number of epochs. 

\subsubsection{Transformer-based methods}

We choose two state-of-the-art transformer-based \cite{Vaswani:Attention} methods, namely Bidirectional Encoder Representations from Transformers (BERT) \cite{devlin-etal-2019-bert} and 
Robustly Optimized BERT Approach (RoBERTa) \cite{Liu:Roberta}. We tested these models with and without data augmentation. 

\subsubsection{Large Language Models (LLMs)}

We selected four state-of-the-art LLMs, namely, OpenAI GPT-4o-mini \cite{Brown2020gpt3, Ouyang2022instructgpt}, Google Gemini-1.5-Pro \cite{geminiteam2023gemini, Imran2024}, LLaMA 3.1 (through Ollama) \cite{touvron:2023:llama, llama3} and Anthropic Claude-3.5-Haiku \cite{claude3}. We evaluated these LLMs in Zero-Shot settings, which means that these LLMs were not trained or fine-tuned on the questions or learning outcomes dataset with annotated labels. Instead, we gave these LLMs a predetermined natural language prompt that told them to designate each sentence with one of the six levels of Bloom. Pydantic AI workflows were used to access OpenAI and Gemini, and custom Python wrappers were used to evaluate LLaMA and Claude. This approach investigated whether modern pretrained LLMs could achieve cognitively-aware classification without being trained, an essential feature of institutions that do not have annotated data to train these data hungry LLMs, nor they have enough resources in the form of compute to train these LLMs.

\section{Experiments and Results} 

The dataset composed of 600 labelled sentences in the same proportion of the six categories of Bloom taxonomy. We did preprocessing by lowercasing, punctuation mark elimination, tokenisation, lemmatisation, and filtering stopwords. There was a minor increment of lexical diversity (10\%), which was accomplished through a small synonym-replacement augmentation. Traditional ML models had Bag-of-Words with optional POS tags whereas RNN models had padded Keras token sequences with optional GloVe embeddings. HuggingFace tokenisers were used to fine-tune transformer models (BERT, RoBERTa). A stratified 80/20 split was done so that all categories were represented in equal proportions.

The conventional ML models were trained with the help of GridSearchCV and cross-validation, whereas RNNs were trained with Adam \cite{kingma2015adam} and early stopping. AdamW \cite{loshchilov2019adamw} with a warm-up schedule was used to fine-tune transformers with 5-10 epochs. Lastly, GPT-4o-mini, Gemini-1.5-pro, Claude 3.5 Haiku and LLaMa3.1 were tested under a purely zero-shot setting with no examples or fine-tuning. Every experiment was done with the help of scikit-learn, TensorFlow/Keras, and HuggingFace on an i7 CPU, 32GB RAM, and an RTX 3060 graphics card workstation. The accuracy, precision, recall, and micro/macro-f1 were used as the indicators of model performance.

\subsection{Results}

This section presents the evaluation outcomes for all modelling approaches explored in this study, namely traditional machine-learning models, recurrent neural networks (RNNs), transformer-based models, and zero-shot large language models (LLMs). Results are organised thematically, with tables and figures positioned immediately after their corresponding discussion to maximise clarity and readability.

\subsubsection{Machine-Learning Models}

The results of the classical ML methods are given in Table~\ref{tab:traditional_ml}. The classical ML models showed good and robust results in the classification of questions and learning outcomes based on Bloom taxonomy. Among all supervised models, Support Vector Machines (SVM) had the highest accuracy (.94) with synonym-based data augmentation. Augmentation also helped significantly to Logistic Regression and raised the accuracy of the model, which was 0.78 to 0.91. Naive Bayes (NB) performed averagely though it recorded a significant improvement when augmented. The POS-tag features had both positive and negative impacts since sometimes they enhanced the accuracy of the features, but in some cases, they also diminished the recall. There was a low degree of overfitting in SVM with almost the same levels of training and validation, but NB and LR had training-validation differences of around 0.10. These results, in general, indicate that the simpler ML classifiers, with specific augmentation, can be used to deliver effective and reliable performance using small datasets.

\begin{table*}[]
    \centering
    \caption{Performance of traditional machine-learning models for Bloom's Taxonomy classification.}
    \label{tab:traditional_ml}
    \scriptsize
    \setlength{\tabcolsep}{2.2pt}
    \renewcommand{\arraystretch}{1.05}
    \begin{tabularx}{\columnwidth}{l C C C C C}
        \toprule
        \textbf{Model} & \textbf{Accuracy} & \textbf{Precision} & \textbf{Recall} & \textbf{F1\textsubscript{micro}} & \textbf{F1\textsubscript{macro}} \\
        \midrule
        Na\"ive Bayes (NB)            & 0.74 & 0.74 & 0.74 & 0.74 & 0.74 \\
        NB w/ POS                     & 0.70 & 0.70 & 0.70 & 0.70 & 0.70 \\
        NB w/ Augmentation            & 0.85 & 0.86 & 0.85 & 0.85 & 0.85 \\
        Logistic Regression (LR)      & 0.78 & 0.78 & 0.78 & 0.78 & 0.78 \\
        LR w/ POS                     & 0.74 & 0.75 & 0.74 & 0.74 & 0.74 \\
        LR w/ Augmentation            & 0.91 & 0.92 & 0.91 & 0.91 & 0.91 \\
        Support Vector Machine (SVM)  & 0.75 & 0.75 & 0.75 & 0.75 & 0.75 \\
        SVM w/ POS                    & 0.74 & 0.78 & 0.74 & 0.74 & 0.75 \\
        \textbf{SVM w/ Augmentation}  & \textbf{0.94} & \textbf{0.95} & \textbf{0.94} & \textbf{0.94} & \textbf{0.94} \\
        \bottomrule
    \end{tabularx}
\end{table*}

\subsubsection{Recurrent Neural Network Models}
The results of the RNN models are given in Table~\ref{tab:rnn_results}. The models based on RNN, such as LSTM, BiLSTM, GRU, and BiGRU, were very prone to overfitting. Baseline LSTM performance was 0.57-0.64 accuracy, though with somewhat higher numbers at 0.63 of GRU. The addition of GloVe embeddings led to better semantic understanding, whereas the addition of augmentation gave moderate increases in accuracy. The most successful variants of RNNs (LSTM with 0.73 accuracy augmentation, GRU with augmentation + embeddings with 0.71 accuracy), however, still had large training validation gaps. Unidirectional architectures (BiLSTM and BiGRU) did not outperform unidirectional ones even with a larger model capacity. The findings of this research show that deep recurrent networks are likely to overfit when working on small-size datasets. 

\begin{table*}[]
    \centering
    \caption{Performance of RNN-based models for Bloom's Taxonomy classification.}
    \label{tab:rnn_results}
    \scriptsize
    \setlength{\tabcolsep}{2.2pt}
    \renewcommand{\arraystretch}{1.05}
    \begin{tabularx}{\columnwidth}{l C C C C C}
        \toprule
        \textbf{Model} & \textbf{Accuracy} & \textbf{Precision} & \textbf{Recall} & \textbf{F1\textsubscript{micro}} & \textbf{F1\textsubscript{macro}} \\
        \midrule
        LSTM                     & 0.64 & 0.65 & 0.64 & 0.64 & 0.62 \\
        LSTM w/ Augmentation     & \textbf{0.73} & \textbf{0.74} & \textbf{0.73} & \textbf{0.73} & \textbf{0.72} \\
        LSTM w/ GloVe            & 0.69 & 0.71 & 0.69 & 0.69 & 0.69 \\
        LSTM w/ GloVe + Augmentation     & 0.72 & 0.72 & 0.72 & 0.72 & 0.72 \\
        BiLSTM                   & 0.57 & 0.57 & 0.57 & 0.57 & 0.56 \\
        BiLSTM w/ Augmentation   & 0.59 & 0.59 & 0.59 & 0.59 & 0.58 \\
        BiLSTM w/ GloVe          & 0.59 & 0.59 & 0.59 & 0.59 & 0.58 \\
        BiLSTM w/ GloVe + Augmentation   & 0.61 & 0.66 & 0.61 & 0.61 & 0.62 \\
        GRU                      & 0.63 & 0.64 & 0.63 & 0.63 & 0.61 \\
        GRU w/ Augmentation      & 0.69 & 0.72 & 0.69 & 0.69 & 0.69 \\
        GRU w/ GloVe             & 0.69 & 0.70 & 0.69 & 0.69 & 0.68 \\
        GRU w/ GloVe + Augmentation      & 0.71 & 0.72 & 0.71 & 0.71 & 0.71 \\
        BiGRU                    & 0.58 & 0.59 & 0.58 & 0.58 & 0.58 \\
        BiGRU w/ Augmentation    & 0.62 & 0.64 & 0.62 & 0.62 & 0.62 \\
        BiGRU w/ GloVe           & 0.59 & 0.61 & 0.59 & 0.59 & 0.57 \\
        BiGRU w/ GloVe + Augmentation    & 0.62 & 0.64 & 0.62 & 0.62 & 0.62 \\
        \bottomrule
    \end{tabularx}
\end{table*}

\subsubsection{Transformer-based Models}
\begin{figure}[ht]
    \centering
    \includegraphics[width=1\linewidth]{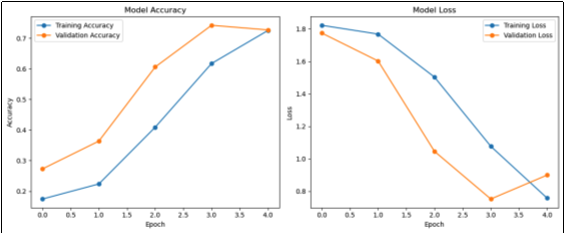}
    \caption{RoBERTa training and validation accuracy/loss curves.}
    \label{fig:roberta_training_curves}
\end{figure}

The results of the Transformer-based models, namely, BERT and RoBERTa are given in Table~\ref{tab:transformer_results}. Models that were based on transformers showed a high performance divide. BERT was overfit and its accuracy was poor (0.35 without augmentation; 0.47 with augmentation). On the other hand, RoBERTa was much more accurate with a score of approximately 0.83. It is probably improved by the fact that RoBERTa uses a more optimised pretraining process and has strong tokenisation. However, the validation loss curve of RoBERTa (see Figure \ref{fig:roberta_training_curves}) shows a definite indication of overfitting towards the final epochs, indicating that transformer fine-tuning remains a difficult task to perform with small datasets.

\begin{table*}[]
    \centering
    \caption{Performance of Transformer-based models (BERT and RoBERTa) for Bloom's taxonomy classification.}
    \label{tab:transformer_results}
    \scriptsize
    \setlength{\tabcolsep}{2.2pt}
    \renewcommand{\arraystretch}{1.05}
    \begin{tabularx}{\columnwidth}{l C C C C C}
        \toprule
        \textbf{Model} & \textbf{Accuracy} & \textbf{Precision} & \textbf{Recall} & \textbf{F1\textsubscript{micro}} & \textbf{F1\textsubscript{macro}} \\
        \midrule
        BERT & 0.35 & 0.45 & 0.35 & 0.35 & 0.35  \\
        BERT + Augmentation & 0.47 & 0.50 & 0.47 & 0.47 & 0.46 \\
        RoBERTa & 0.78 & 0.78 & 0.78 & 0.78 & 0.78 \\
        RoBERTa + Augmentation & 0.83 & 0.86 & 0.83 & 0.83 & 0.83 \\
        \bottomrule
    \end{tabularx}
\end{table*}

\subsubsection{Large Language Models (LLMs)}
The results of the applying LLMs in zero-shot settings are given in Table~\ref{tab:llm_results}. The results of zero-shot classification with the use of LLMs suggest that the current pretrained models can perform moderately to strongly without any supervised training. Google Gemini-1.5-pro and OpenAI GPT-4o-mini reached the values of 0.72 and 0.73 respectively. Anthropic Claude-3.5-haiku had a score of 0.58 and LLaMA3.1 (through Ollama) was the weakest with a score of 0.42. These findings underscore the fact that, in the context where laboratory-labelled data is scarce or not accessible, LLMs still struggle to reach classical ML methods.

\begin{table*}[]
    \centering
    \caption{Zero-shot performance of Large Language Models (LLMs).}
    \label{tab:llm_results}
    \scriptsize
    \setlength{\tabcolsep}{2.2pt}
    \renewcommand{\arraystretch}{1.05}
    \begin{tabularx}{\columnwidth}{l C C C C C}
        \toprule
        \textbf{Model} & \textbf{Accuracy} & \textbf{Precision} & \textbf{Recall} & \textbf{F1\textsubscript{micro}} & \textbf{F1\textsubscript{macro}} \\
        \midrule
        OpenAI GPT-4o-mini     & 0.72 & 0.76 & 0.72 & 0.72 & 0.72 \\
        Google Gemini-1.5-pro  & 0.73 & 0.77 & 0.73 & 0.74 & 0.73 \\
        Ollama LLaMA3.1        & 0.42 & 0.62 & 0.42 & 0.42 & 0.35 \\
        Claude 3.5 Haiku       & 0.58 & 0.64 & 0.58 & 0.58 & 0.51 \\
        \bottomrule
    \end{tabularx}
\end{table*}

\section{Discussion}

One of the issues that was evident in all the experiments was high risk of over-fitting by large number of models. The number of samples per Bloom category was limited to 100, which is insufficient to run many deep learning models. This restriction was also manifested by the fact that the training accuracies were high (all of them were above 0.95) whereas the validation scores are significantly smaller (around 0.80-0.90). Such gaps can be small but when the number of data is limited, the difference of $0.10$ to 0.11 is enormous and reflects memorisation as opposed to generalisation.

Data augmentation by means of synonyms was introduced and helped reduce overfitting to a certain degree, as it added lexical diversity but did not change the meaning of the sentences. This methodology brought significant gains, of 5\% to 10\% accuracy and F1-scores, to traditional ML architectures as well as to RNN architectures. Nonetheless, synonym replacement does not occur without its negative aspects: badly paired synonyms may cause semantic drift, sometimes changing the cognitive intent of a question. In spite of those restrictions, embedding-based modifications like GloVe vectors also enhanced the results of RNNs by providing better semantic structure. Nonetheless, the results of RNN models remained inferior to the most successful traditional ML solution, which again highlights the fact that even deeper architectures may not be beneficial in cases where data size is limited.

Transformer models had kinds of mixed behaviour. Compared to BERT, which had serious issues with overfitting, presumably because it has a large number of parameters and is sensitive to small training corpora, RoBERTa was found to be more resilient. This could be due to its stronger pretraining goals and its tokenisation policies that enable it to learn linguistic details on fewer samples. However, RoBERTa also ultimately exhibited overfitting in subsequent epochs, which underscores the necessity of aggressive regularisation or early termination when using transformers on small learning data.

A very interesting comparison was the zero-shot assessment of large language models (LLMs). With no finetuning, OpenAI models like GPT (gpt-4o-mini) and Google models like Gemini reached moderate accuracy in the range of $0.72$ to $0.73$. The given performance is more impressive since there is no task-specific training at all, which implies that the performance of LLMs is highly innate with respect to the ability to reason at a cognitive level by virtue of their wide-range pretraining corpora. Conversely, LLaMA (through Ollama) and Claude were more prompt-sensitive and less accurate, which indicates differences in architectural and training in the families of LLM. Such results suggest that mindful prompt engineering or chain-of-thought prompting may be used to further boost the performance of LLM in the zero-shot setting.

\begin{figure}
    \centering
    \includegraphics[width=\linewidth]{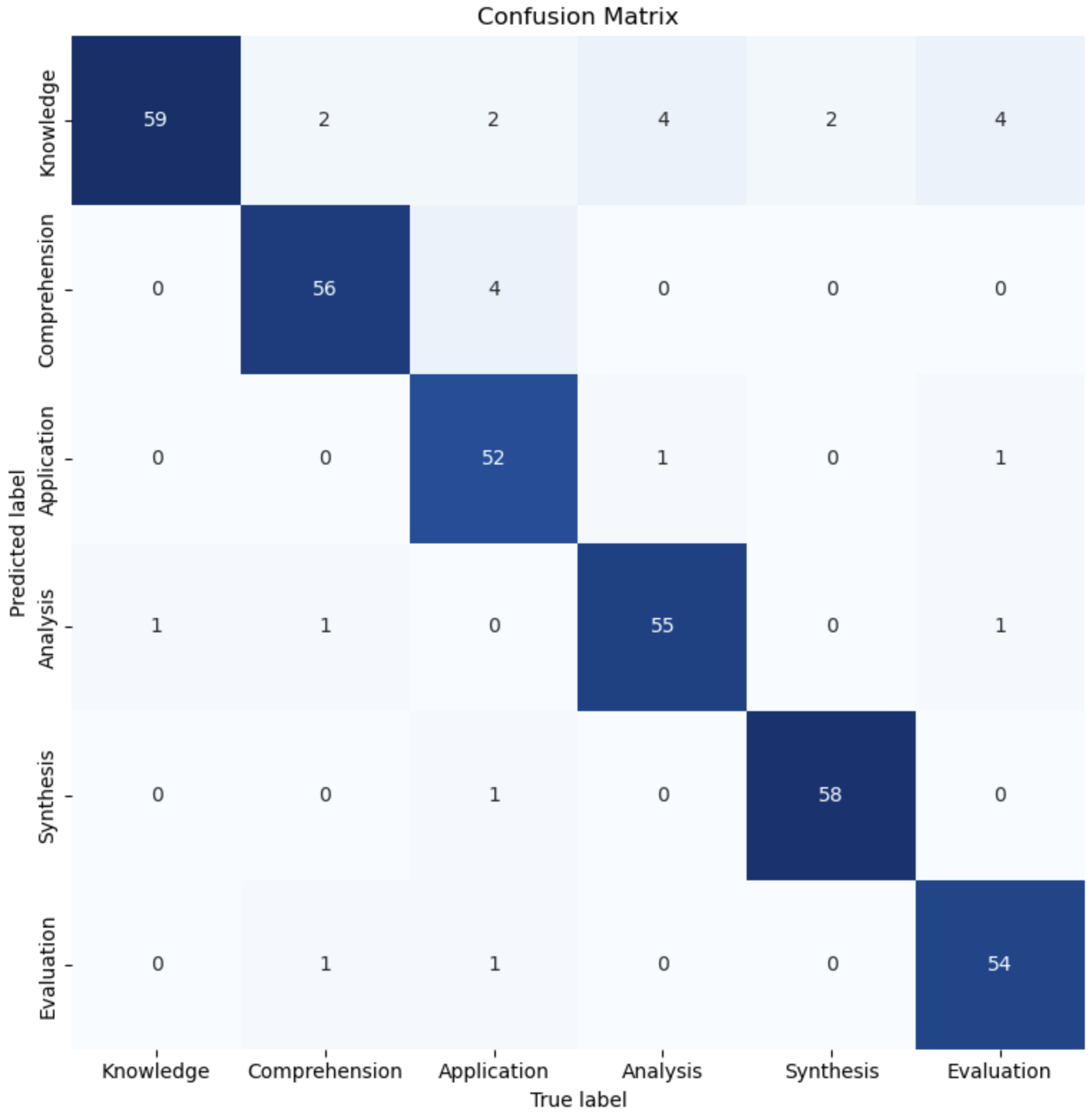}
    \caption{Confusion Matrix for classifying questions into six categories according to Bloom's taxonomy by SVM.}
    \label{fig:SVM_confusion_matrix}
\end{figure}

In general, the findings validate that it is possible to automate Bloom taxonomy classification, though, only by means of proper preprocessing, selected augmentation, and selection of models. The best performance (94 percent accuracy/F1), and least overfitted, was obtained with traditional ML methods, in particular, SVM using augmentation. The confusion matrix for classifying questions into six categories according to Bloom's taxonomy by SVM is given in Figure~\ref{fig:SVM_confusion_matrix}. Most of the misclassifications happens at the adjacent cognitive levels, indicating that there are subtle differences and it's hard to come up with discriminative features for separating classes with 100\% accuracy. The small dataset size impacted negatively on RNNs and BERT whereas RoBERTa did not disappoint but was prone to overfitting. The zero-shot capabilities of LLM were also reported to be strong and provide an effective alternative in situations where annotated data or computational resources are scarce.

Although the results are promising, there are limitations of our study. First, we work on a small-size and a single dataset. Second, we applied basic text augmentation techniques. The limitations of this work can be improved by:
\begin{itemize}
    \item gathering more and larger labelled data on categories of Bloom,
    \item researching the more complex augmentation strategies, including paraphrasing, back-translation and generative data synthesis,
    \item research on prompt engineering or parameter-efficient fine-tuning (e.g., LoRA \cite{LoRA}) of LLMs,
    \item using interpretability methods (e.g. SHAP \cite{SHAP}, attention visualisation) to learn decision processes in both classical and neural networks.
\end{itemize}

These directions will improve the strength, interpretability, and a practical implementation of automated systems of the Bloom classification of taxonomy in the real-life context of education.

\section{Conclusion}
This paper studied the possibility of automating the classification of questions or learning outcomes based on Bloom taxonomy with a variety of different NLP methods, including standard machine-learning models, deep neural networks, transformer models, and large language models. Although the methods are diverse, the results are consistently the same in that the choice of model has to be affected by the magnitude and character of the available data. Traditional models, in particular, SVM with synonym based augmentation had the most reliable and overall best results, with 94\% accuracy with insignificant overfitting. Conversely, RNN-based architectures and BERT had a hard time with generalisation, demonstrating the inability to train the deep models with small educational datasets. RoBERTa was significantly more effective, and it was sensitive to the lack of data.

The zero-shot test of LLams also was an encouraging alternative, with OpenAI and Gemini getting an approximation of 0.72-0.73 without any task-specific training. These findings imply that massive pretrained models will be useful in real educational environments with constrained labelled data or computes.
However, collectively, it can be seen that automated Bloom-level classification is possible, although the success hinges on the ability to match the complexity of the model with the volume of the dataset and the ability to provide the support system with the necessary augmentation techniques. Future research work will focus on increasing the size of annotated datasets, more sophisticated augmentation and prompt-engineering, and implement interpretability technologies to understand how models can differentiate between cognitive levels. Further development of such directions will assist in converting automated classification of Taxonomy of Bloom into de facto scalable assessment design and learning analytics tools.

\section*{Acknowledgment}
This research was supported by Katana, the high performance computing facility at the University of New South Wales. The authors also  acknowledge the financial support provided by the School of Computer Science and Engineering, Faculty of Engineering, UNSW Sydney. 

\bibliographystyle{IEEEtran}
\bibliography{references} 

\end{document}